\documentclass[10pt,letterpaper]{article}

\usepackage{cogsci}

\cogscifinalcopy 

\usepackage[
  style=apa,
  natbib=true,
  annotation=false,
]{biblatex}
\usepackage{float} 

\usepackage{graphicx}  
\usepackage{booktabs}  
\usepackage[table]{xcolor}  
\usepackage[most]{tcolorbox}

\usepackage{subcaption}
\usepackage{tabularx}
\usepackage{array}
\usepackage{subcaption}
\title{Dynamic Theory of Mind as a Temporal Memory Problem: Evidence from Large Language Models}

\author[1]{\mbox{Thuy Ngoc Nguyen}}
\author[2]{\mbox{Duy Nhat Phan}}
\author[3]{\mbox{Cleotilde Gonzalez}}
\affil[1]{Department of Computer Science, University of Dayton}
\affil[2]{University of Dayton Research Institute, University of Dayton}
\affil[3]{Department of Social and Decision Sciences, Carnegie Mellon University}

\begin{document}

\maketitle

\begin{abstract}
Theory of Mind (ToM) is central to social cognition and human–AI interaction, and Large Language Models (LLMs) have been used to help understand and represent ToM. However, most evaluations treat ToM as a static judgment at a single moment, primarily relying on tests of false beliefs. This overlooks a key dynamic dimension of ToM: the ability to represent, update, and retrieve others' beliefs over time. We investigate dynamic ToM as a temporally extended representational memory problem, asking whether LLMs can track belief trajectories across interactions rather than only inferring current beliefs. We introduce DToM-Track, an evaluation framework to investigate temporal belief reasoning in controlled multi-turn conversations, testing the recall of beliefs held prior to an update, the inference of current beliefs, and the detection of belief change. 
Using LLMs as computational probes, we find a consistent asymmetry: models reliably infer an agent's current belief but struggle to maintain and retrieve prior belief states once updates occur. This pattern persists across LLM model families and scales, and is consistent with recency bias and interference effects well documented in cognitive science.
These results suggest that tracking belief trajectories over time poses a distinct challenge beyond classical false-belief reasoning. By framing ToM as a problem of temporal representation and retrieval, this work connects ToM to core cognitive mechanisms of memory and interference and exposes the implications for LLM models of social reasoning in extended human-AI interactions.

\textbf{Keywords:}
Theory of mind; temporal belief tracking; large language models; social cognition; human–AI interaction.
\end{abstract}

\section{Introduction}

Theory of Mind (ToM) refers to the capacity to represent and reason about others' unobserved mental states, such as beliefs, intentions, and desires, as internal representations that guide social interaction \citep{premack1978}. While ToM is often studied as belief attribution at a single moment, social reasoning in natural interaction unfolds over time: agents must infer current beliefs while maintaining and retrieving earlier belief representations as new information is introduced. This temporally extended aspect of ToM is essential for everyday conversation, where beliefs are revised, corrected, or displaced over time, and is increasingly critical in human–AI interaction, where systems must adapt to users' evolving goals and beliefs~\citep{walsh2025theory,lee2020perceiving}.

Cognitive science research suggests that the success of such temporal belief tracking is shaped by general-purpose memory and judgment mechanisms that operate over time~\citep{hogarth1992order}. In particular, belief reasoning is systematically influenced by recency bias, whereby recent information disproportionately affects judgments~\citep{murdock1962serial}, as well as by interference effects, in which updated representations compete with and disrupt access to earlier belief states~\citep{hoch1984availability, birch2007curse}. 
These mechanisms imply that retrieving prior beliefs after an update may pose a distinct cognitive challenge, even when current belief attribution remains accurate. From this perspective, ToM is not only a matter of representational competence but also of maintaining and retrieving mental state representations under temporal and memory constraints.

Despite evidence that belief reasoning over time is shaped by recency and interference, most of the ToM literature, both in cognitive science and in computational modeling, has focused on static belief attribution~\citep{hu2025re}. Canonical tasks typically ask what an agent believes at a single moment (e.g.,``What does X believe?''), often contrasting that belief with reality in false belief paradigms \citep{wimmer1983beliefs,le2019revisiting}. While such tasks have been instrumental in establishing the representational nature of ToM, they treat belief attribution as a snapshot inference rather than as a process unfolding over interaction. As a result, it remains underspecified how mental state representations are maintained, updated, and retrieved over time, and how memory constraints affect access to earlier beliefs once new information is introduced. Moreover, this emphasis on false belief has narrowed the scope of ToM evaluation, despite evidence that social reasoning involves a broader range of mental states, including intentions, desires, emotions, and knowledge~\citep{ma2023towards}.

\begin{figure*}[t]
\centering
\begin{subfigure}[t]{1\linewidth}
  \centering
  \includegraphics[width=0.9\linewidth]{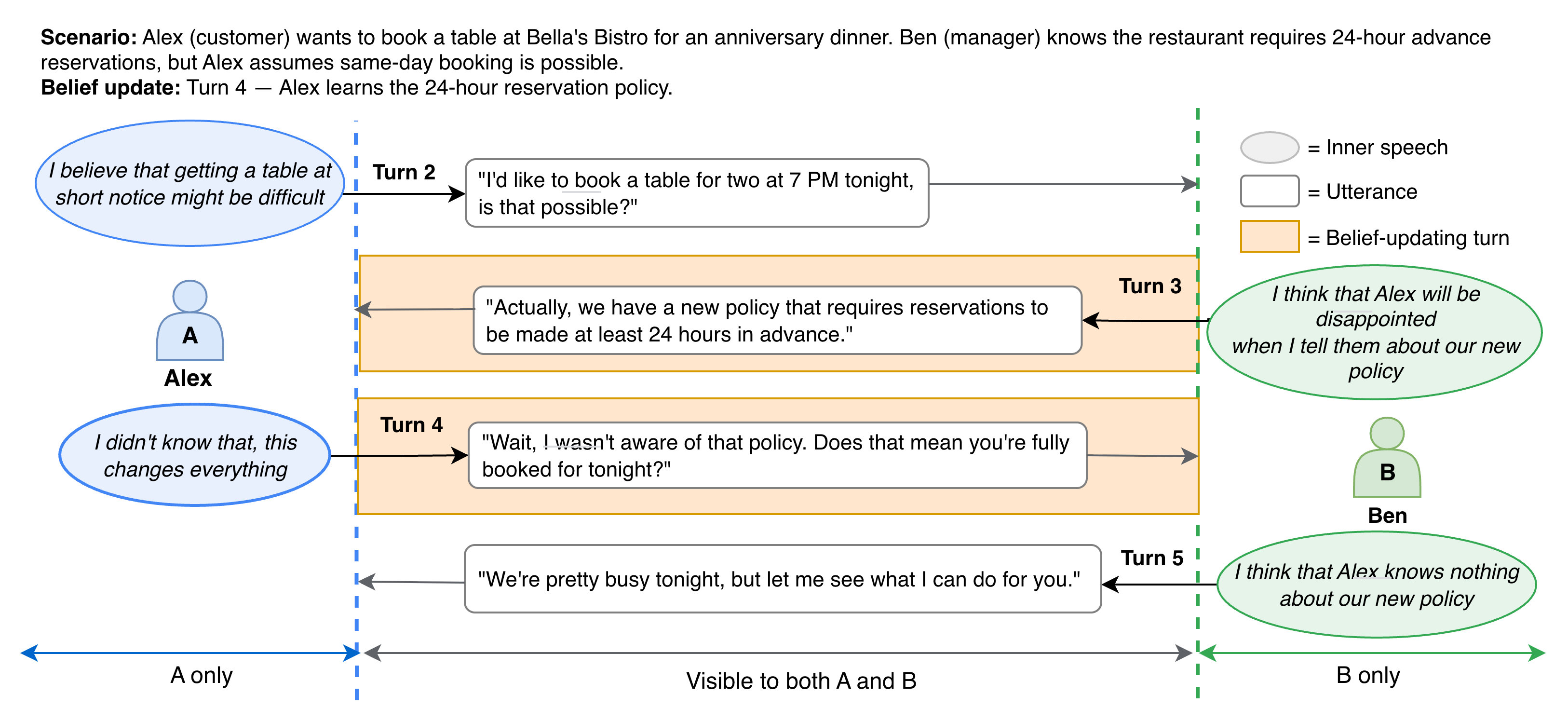}
  \caption{Example interaction with hidden inner speech and planned belief updates.}
  \label{fig:dtom_example_a}
\end{subfigure}

\vspace{1em}

\begin{subfigure}[t]{0.9\linewidth}
\centering
\small
\setlength{\tabcolsep}{6pt}
\renewcommand{\arraystretch}{1.08}

\begin{tabular}{@{}p{0.32\linewidth}p{0.32\linewidth}p{0.32\linewidth}@{}}
\toprule
\textbf{Pre-Update (prior belief)} &
\textbf{Post-Update (current belief)} &
\textbf{Update Detection} \\
\midrule
\textit{Before Turn 3, what did Alex believe about reservations?} &
\textit{After Turn 4, what does Alex believe about reservations?} &
\textit{At which turn did Alex’s belief change?} \\
\addlinespace[0.25em]

\begin{tabular}[t]{@{}l@{}}
\textcolor{gray}{(A) 24-hour notice required}\\
\textcolor{gray}{(B) Online booking only}\\
\textcolor{gray}{(C) No reservations needed}\\
\textbf{(D) Same-day accepted} \checkmark

\end{tabular}
&
\begin{tabular}[t]{@{}l@{}}
\textcolor{gray}{(A) Same-day accepted}\\
\textcolor{gray}{(B) No reservations needed}\\
\textbf{(C) 24-hour notice required} \checkmark\\
\textcolor{gray}{(D) Online booking only}
\end{tabular}
&
\begin{tabular}[t]{@{}l@{}}
\textcolor{gray}{(A) Turn 2}\\ \textbf{(B) Turn 3--4} \checkmark\\
\textcolor{gray}{(C) Turn 5}\\ \textcolor{gray}{(D) Did not change}
\end{tabular}
\\
\bottomrule
\end{tabular}

\caption{Illustrative temporal question types evaluating prior belief, current belief, and belief-change timing.}
\label{fig:dtom_example_b}
\end{subfigure}

\caption{Illustration of DToM-Track.
(a) Role-playing interaction with information asymmetry induced by hidden inner speech and mid-conversation belief updates.
(b) Examples of temporal questions examining belief recall and change.}
\label{fig:dtom_example}
\end{figure*}

Computational cognitive approaches that use artificial agents to test hypotheses about the structure and limitations of human social reasoning provide an initial means to isolate temporal and memory components of ToM that are difficult to observe in static evaluations~\citep{baker2011bayesian,nguyen2022theory,kostka2025towards}. However, many existing approaches model belief attribution as inference over a limited state space, emphasizing belief updating while underemphasizing belief maintenance, retrieval after revision, and interference over extended interaction.
Contemporary large language models (LLMs) can successfully infer others' mental states in constrained one-shot settings~\citep{brown2020language,kosinski2023theory}, largely by encoding broad patterns of human behavior from large scale training data. However, their performance varies widely across tasks and formulations~\citep{strachan2024testing,street2025llms,ullman2023large,shapira2024clever}. We adopt the view that LLMs can serve as computational probes for examining how belief representations are constructed and accessed over time. In particular, evaluating whether models can retrieve beliefs held prior to an update, rather than only inferring the current belief state, isolates temporal and memory-related components of ToM that are unclear in static evaluations.


To operationalize this temporally extended view of ToM, we introduce DToM-Track, an evaluation framework designed to probe how mental state representations are maintained, updated, and retrieved across interaction (Fig.~\ref{fig:dtom_example}). DToM-Track tests whether LLMs can track belief trajectories over the course of a conversation, including recalling beliefs held prior to a change, inferring beliefs after an update, and identifying when belief revisions occur. The framework adopts a controlled generation paradigm based on inner-speech prompting~\citep{tomato2024}, in which agents verbalize their mental states before each utterance while these internal representations remain concealed from their conversational partners. This induced information asymmetry~\citep{guo2023suspicion} mirrors false belief settings~\citep{wimmer1983beliefs,brauner2020being} and enables systematic evaluation of mental states. By maintaining structured mental states across turns, DToM-Track isolates temporal and memory-dependent components of ToM that are absent from static evaluations.

Using controlled LLM–LLM conversations, DToM-Track introduces temporal question types instantiated via structured templates that directly test belief dynamics, including pre-update, post-update, and update-detection questions. Together with standard temporal, second-order, and false belief questions, they assess ToM across multiple mental states, including beliefs, intentions, desires, emotions, and knowledge.

Applying DToM-Track to six LLMs as computational probes reveals a consistent dissociation in dynamic belief reasoning. Models reliably infer an agent's current-belief but perform substantially worse when recalling beliefs held prior to an update. This pattern is consistent with recency bias and interference, with recent updates dominating access to earlier states. This difficulty exceeds that observed in standard ToM tasks such as false belief reasoning and persists with increasing model scale, indicating a limitation in temporal representation and retrieval rather than model capacity. Together, these findings identify dynamic belief tracking as a distinct component of ToM shaped by memory and interference effects.


\begin{figure*}[t]
\centering
\includegraphics[width=\linewidth]{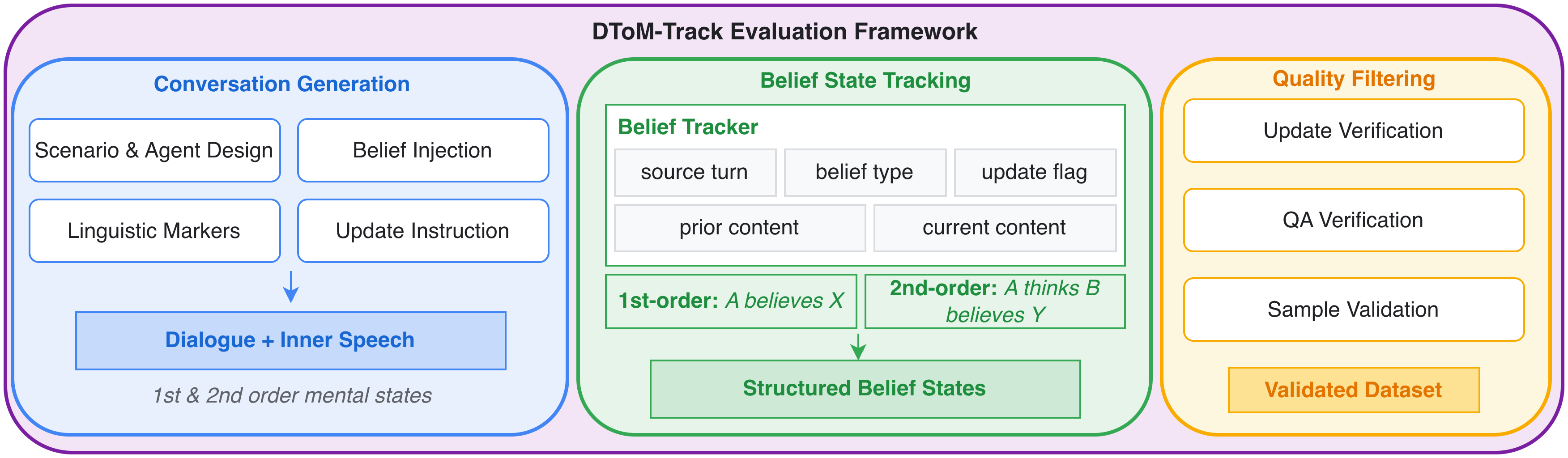}
\caption{DToM-Track framework. Controlled conversation generation produces dialogues with inner speech and planned belief updates; a belief tracker maintains structured first- and second-order mental states across turns; and an LLM-based filtering pipeline verifies update realization and question answerability for dynamic ToM evaluation.}
\label{fig:dtom_framework}
\end{figure*}

\section{Method}
Several frameworks evaluate ToM in LLMs, including Hi-ToM~\citep{hitom2023}, SOTOPIA~\citep{zhou2023sotopia}, OpenToM~\citep{opentom2024}, TomBench~\citep{tombench2024}, and ToMATO~\citep{tomato2024}, showing that LLMs can reason about others' mental states, including recursive beliefs of moderate complexity~\citep{jones2024}. Recent work has introduced temporal structure into ToM evaluation, such as belief dynamics annotation~\citep{minddial2024} and conversational adaptability studies of intent tracking~\citep{conv_adapt2025}. However, these approaches emphasize belief updating at each turn and have yet to study whether models can maintain and query belief trajectories across interactions, such as recalling beliefs held prior to a correction or capturing cognitive biases in temporal belief tracking.

DToM-Track addresses this gap by building on prior work that uses LLMs to generate conversational data~\citep{kim2023fantom,tomato2024,zhou2023sotopia}. Specifically, it employs controlled LLM–LLM interactions to construct multi-turn conversations with planned belief updates. Unlike prior work that emphasizes static mental-state inference, DToM-Track evaluates whether models can track belief changes over interactions, including maintaining and recalling beliefs held prior to an update rather than only representing the current belief state. Following prior work, we use LLaMA-3-70B-Instruct~\citep{dubey2024llama} for LLM–LLM conversation generation due to its transparency and strong performance~\citep{chiang2024chatbot}.
The framework comprises three components: controlled conversation generation, belief state tracking, and quality filtering (Fig.~\ref{fig:dtom_framework}).


\subsection{Conversation Generation}
Building on inner-speech prompting~\citep{tomato2024}, DToM-Track generates controlled multi-turn dialogues in which belief updates are explicitly planned at predefined turns, rather than inferred post hoc. Specifically, dialogues are produced via LLM–LLM interaction, with each agent’s mental state tracked through inner-speech annotations. Each conversational turn is formatted as:
\begin{quote}
\texttt{Agent: (thought) ``utterance''}
\end{quote}
\noindent where the \textit{}{thought} represents the agent's current mental state and the \textit{utterance} is the spoken dialogue. Each mental state type uses a specific inner speech prefix to guide generation:
\begin{quote}\small
\texttt{Belief: (I believe that \ldots); Intention: (I will \ldots); Desire: (I want \ldots); Emotion: (I feel \ldots); Knowledge: (I know \ldots)}
\end{quote}


\begin{table}[t]
\centering
\caption{Scenario generation prompt (abbreviated) with planned belief changes.}
\label{tab:scenario_prompt}
\begin{tcolorbox}[
  colback=gray!5,
  colframe=gray!60,
  width=\columnwidth,
  boxrule=0.3mm,
  arc=2mm,
  top=4pt,
  bottom=4pt,
  left=4pt,
  right=4pt,
  before skip=0pt,  
  after skip=0pt  
]
\small
\texttt{Generate a realistic conversation scenario for testing Theory of Mind and dynamic belief tracking.}
\smallskip
\textbf{Domain:} \texttt{\{domain\}} 

\smallskip
\textbf{Conversation Structure:}\\
\texttt{Turn 0--1: Greetings} \\
\texttt{Turn 2--3: Establish context} \\
\texttt{Turn 4--8: Main conversation} \\
\texttt{Turn 9: Final turn}

\smallskip
\textbf{Turn-taking:} \texttt{Agents alternate (A: even; B: odd). Plan belief updates at turn 4 or later.}

\smallskip
\textbf{Requirements:}\\
\texttt{1. Create two agents with different roles.}\\
\texttt{2. Each agent has private information unknown to the other.}\\
\texttt{3. Scenario must include at least \{min\_belief\_changes\} belief changes.}\\
\texttt{4. Include belief updates via correction, learning, or goal change.}

\smallskip
\textbf{Output Format:}\\
\texttt{\{scenario\_id, domain, context,}
\texttt{agent\_a: \{name, role, private\_info, initial\_belief, goal\},}
\texttt{agent\_b: \{...\},}
\texttt{planned\_belief\_changes: [\{turn, agent, trigger\_type, old\_belief, new\_belief\}]\}}

\end{tcolorbox}
\end{table}

Importantly, inner-speech annotations are hidden from the conversational partner, inducing information asymmetry that mirrors human interaction. This asymmetry ensures agents have only partial access to others' mental states, enabling principled evaluation of false beliefs~\citep{brauner2020being}.

\noindent \textit{\textbf{Scenario and Agent Design.}}
 Each conversation is grounded in a scenario specifying (a) a conversational domain (e.g., restaurant booking, travel planning, medical consultation), (b) two agents with distinct roles, goals, and private information, and (c) planned belief updates at designated turns, typically between turns 4 and 8. Agents are also assigned big five personality traits to introduce natural variation in conversational style~\citep{de2000big,zhou2023sotopia}. The scenario generation prompt (Table~\ref{tab:scenario_prompt}) instructs an LLM to produce scenarios with explicitly planned belief changes. Each agent receives a system prompt defining its role, personality, private information, and goal, and is instructed to verbalize its mental states as inner speech prior to each utterance.

\noindent \textit{\textbf{Controlled Belief Injection.}}
DToM-Track explicitly plans belief updates at predetermined conversational turns prior to generation. This design ensures that belief changes occur at known points in the interaction, enabling evaluation of temporal belief tracking. By controlling when updates happen, the framework supports queries such as ``What did X believe before turn 5?'' and allows direct verification of whether models can maintain and retrieve prior belief states rather than relying on post hoc inference from unconstrained dialogue.

\noindent \textit{\textbf{Linguistic Update Markers.}}
Belief updates are signaled through linguistic markers adapted from prior work on conversational adaptability~\citep{conv_adapt2025}. Specifically, we employ 10 implicit correction templates (e.g., ``Actually,...'', ``I just realized...'', ``Wait,...'') and 5 explicit negation patterns (e.g., ``I mistakenly said X, but it's actually Y''). These markers provide naturalistic cues for belief change.
Importantly, models are not instructed to use these markers during evaluation. They appear in the dialogue context to support natural conversational flow. We leave explicit manipulation of the markers for future work.

\noindent \textit{\textbf{Belief Update Instruction.}}
At designated update turns, agents are given explicit instructions defining a belief transition, including the prior belief, the updated belief, and the change type (e.g., correction, new information, contradiction, or goal shift). Agents are instructed to express the updated belief in their inner speech and to signal the change using a prescribed linguistic marker in their utterance. This controlled procedure allows belief updates to occur at known points in the dialogue while preserving natural conversational flow.

\subsection{Belief State Tracking}
DToM-Track maintains structured belief representations across turns rather than treating mental states independently at each utterance. For each belief, the tracker records (i) the \textit{source turn} in which it was expressed, (ii) the \textit{belief type} (belief, intention, desire, emotion, or knowledge), (iii) an \textit{update flag} indicating whether the belief has been superseded, and (iv) the \textit{prior content} before any update. This representation supports systematic generation of temporal queries that distinguish prior from current belief states.

DToM-Track extracts second-order beliefs (beliefs about another agent's beliefs) via pattern matching over inner-speech annotations, using prefixes such as \texttt{(I think that \{other\} believes …)}. This supports construction of second-order and false belief questions.


\subsection{Quality Filtering}
A key challenge in LLM-based evaluation is ensuring that planned belief updates are expressed in generated conversations and that questions are answerable. Drawing on verifiable generation approaches that use LLMs as automated verifiers~\citep{liu2023evaluating,li2024llatrieval}, we employ a multi-stage LLM-based filtering pipeline:

\noindent \textit{\textbf{Update Verification.}}
For each generated conversation, an LLM-based verifier assesses whether planned belief updates occur as intended. The verifier scans each turn for linguistic cues signaling belief change and aligns detected updates with the predefined update schedule. Conversations are retained only if all planned updates are detected and verification performance exceeds an F1 threshold of 0.5.

\noindent \textit{\textbf{QA Verification.}}
Each generated question–answer item undergoes LLM-based verification along three criteria~\citep{chang2024survey}: (i) \textit{correctness}, whether the labeled answer follows from the conversation; (ii) \textit{soundness}, whether the question is unambiguous and answerable from context alone; and (iii) \textit{quality}, whether distractors are plausible yet clearly incorrect. Items are retained only if all criteria are met and a minimum quality threshold is exceeded.

\noindent \textit{\textbf{Sample Validation.}}
As a final filter, samples are validated by testing whether LLMs produce parseable multiple-choice answers. Items are removed if responses are missing or cannot be mapped to a valid option (A-D), eliminating poorly-formed  or ambiguous questions that could introduce noise.

\begin{table*}[!htpb]
\centering
\small
\caption{Question types in DToM-Track and corresponding templates. 
Variables: \texttt{\{agent\}} = agent name, 
\texttt{\{topic\}} = mental state type, 
\texttt{\{turn\}} = turn number, 
\texttt{\{utterance\}} = quoted speech, 
\texttt{\{other\_agent\}} = conversational partner. 
The first three types (bold) examine dynamic belief tracking.}
\label{tab:dtom_questions}

\setlength{\tabcolsep}{6pt}
\renewcommand{\arraystretch}{1.05}
\resizebox{0.95\linewidth}{!}{%
\begin{tabular}{@{}l p{0.23\linewidth} p{0.58\linewidth}@{}}
\toprule
\textbf{Question Type} & \textbf{What It Tests} & \textbf{Template Examples} \\
\midrule

\textbf{\texttt{pre\_update}} &
What did X believe \textit{before} the change? &
\texttt{What was \{agent\}'s \{topic\} before the update at turn \{turn\}?} \\

\textbf{\texttt{post\_update}} &
What does X believe \textit{after} the change? &
\texttt{After the update at turn \{turn\}, what is \{agent\}'s current \{topic\}?}
\\

\textbf{\texttt{update\_detection}} &
\textit{When} did X’s belief change? &
\texttt{At which point did \{agent\} update their \{topic\}?}
\\

\midrule
\texttt{temporal} &
What did X believe at turn $t$? &
\texttt{At turn \{turn\}, what was \{agent\}'s \{topic\}?}
\\

\texttt{second\_order} &
What does X think Y believes? &
\texttt{At turn \{turn\}, what did \{agent\} think \{other\_agent\} believed about \{topic\}?} \\

\texttt{false\_beliefs} &
What was X’s actual belief and What did Y incorrectly assume? &
\texttt{What did \{agent\} actually believe about \{topic\}, despite \{other\_agent\}'s assumption?}; \texttt{What did \{other\_agent\} incorrectly assume about \{agent\}'s \{topic\}?}  \\

\bottomrule
\end{tabular}}
\end{table*}

\section{Experiments}
Using DToM-Track, we evaluate dynamic ToM in six open-source and proprietary LLMs, assessing belief recall, updating, and change detection across multi-turn interactions, alongside false belief and second-order reasoning tasks.

\noindent \textbf{Dataset.}
We evaluate all LLMs on the finalized DToM-Track dataset, which comprises multi-turn dialogues with explicitly scheduled belief updates, structured mental state annotations, and validated multiple choice questions. Items without detectable belief changes are removed, and all questions are automatically verified for correctness, soundness and quality. The resulting dataset includes 5,794 questions across six question types and five mental state categories. Table~\ref{tab:dataset_stats} reports the distribution by question type and mental state category.

\begin{table}[!htpb]
\centering
\caption{DToM-Track dataset statistics.}
\label{tab:dataset_stats}
\footnotesize

\resizebox{\linewidth}{!}{%
\begin{tabular}{lrr @{\hspace{1cm}} lrr}
\toprule
\multicolumn{3}{c}{\textbf{By Question Type}} &
\multicolumn{3}{c}{\textbf{By Mental State Category}} \\
\cmidrule(lr){1-3}\cmidrule(lr){4-6}

\textbf{Category} & \textbf{Count} & \textbf{\%} &
\textbf{Category} & \textbf{Count} & \textbf{\%} \\
\midrule
Temporal          & 1,807 & 31.2\% & Intention & 1,639 & 28.3\% \\
False Belief      & 1,761 & 30.4\% & Desire    & 1,529 & 26.4\% \\
Second-Order      &   768 & 13.3\% & Belief    & 1,136 & 19.6\% \\
Update Detection  &   591 & 10.2\% & Emotion   &   774 & 13.4\% \\
Post-Update       &   527 &  9.1\% & Knowledge &   716 & 12.4\% \\
Pre-Update        &   340 &  5.9\% &           &       &        \\
\midrule
\textbf{Total} & \textbf{5,794} & \textbf{100\%} &
\textbf{Total} & \textbf{5,794} & \textbf{100\%} \\
\bottomrule
\end{tabular}}
\end{table}


\noindent \textbf{Question Types.}
DToM-Track introduces three question types that evaluate temporal belief tracking, a dynamic dimension of ToM not captured by prior evaluation frameworks such as ToMATO. These questions assess whether models can reason over belief trajectories across interaction: pre-update questions test recall of beliefs held before a change, post-update questions assess beliefs after the change, and update-detection questions identify when a belief update occurred. These temporal questions are instantiated via structured templates and evaluated along with standard temporal, second-order, and false belief questions (Table~\ref{tab:dtom_questions}). 

\noindent \textbf{Evaluated Models.}
We evaluate six LLMs from multiple model families and scales: LLaMA~3.3-70B, Mistral Large, Ministral-14B, GPT-4o-mini, LLaMA~3.1-8B, and LLaMA~3.2-3B. The set includes both open-source (LLaMA, Mistral) and proprietary (GPT-4o-mini) models, spanning 3B–70B parameters and accessed via OpenAI API or AWS Bedrock. All models are evaluated zero-shot using a standardized prompt that presents the conversation context followed by a multiple-choice question. Each model selects one of four answer options, with random baseline accuracy at 25\%.

\begin{table*}[t]
\centering
\caption{Accuracy (\%) is reported by question type. \textbf{Bold} and \underline{underline} indicate the best and second-best performance.}
\label{tab:main_results}
\resizebox{0.94\textwidth}{!}{%
\begin{tabular}{lcccccc}
\toprule
Question Type & GPT-4o-mini & llama3-1-8b & llama3-3-70b & ministral-3-14b & mistral-large-2402 & llama3-2-3b \\
\midrule
Temporal & \underline{57.7} & 35.6 & \textbf{65.5} & 54.4 & 57.4 & 30.2 \\
Update Detection & 75.0 & 47.2 & \underline{76.1} & \textbf{77.8} & 75.1 & 53.6 \\
Post-Update & \underline{68.5} & 57.9 & \textbf{71.3} & 64.9 & 65.7 & 55.0 \\
False Belief & 45.9 & 35.6 & \textbf{59.2} & 42.0 & \underline{51.7} & 34.0 \\
Second-Order & \underline{57.9} & 37.0 & \textbf{62.0} & 43.4 & 49.2 & 34.0 \\
Pre-Update & 27.6 & 12.1 & \textbf{40.9} & 30.0 & \underline{40.0} & 15.6 \\
\midrule
\textbf{Overall} & 55.1 & 37.6 & \textbf{63.3} & 51.1 & \underline{56.1} & 35.7 \\
\bottomrule
\end{tabular}}
\end{table*}

\begin{figure*}[t]
\centering
\begin{subfigure}[t]{0.29\textwidth}
  \centering
  \includegraphics[width=\linewidth]{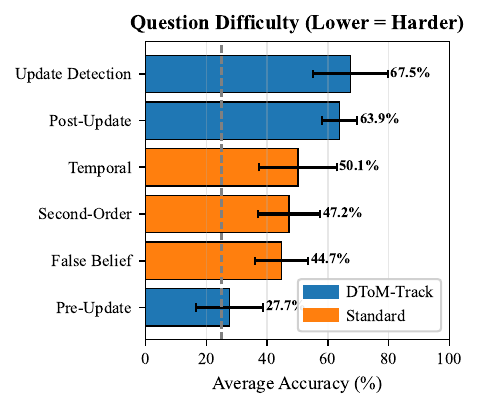}
  \caption{Difficulty by question type (blue bars: DToM-Track temporal questions).}
  \label{fig:question_difficulty}
\end{subfigure}\hspace{0.25em}
\begin{subfigure}[t]{0.35\textwidth}
  \centering
  \includegraphics[width=\linewidth]{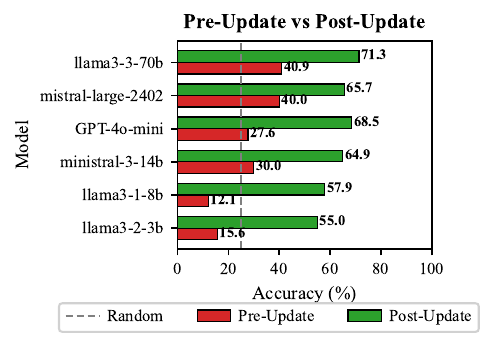}
  \caption{Post-update (current) vs. pre-update (prior) belief performance.}
  \label{fig:recency_bias}
\end{subfigure}\hfill
\begin{subfigure}[t]{0.35\textwidth}
  \centering
  \includegraphics[width=\linewidth]{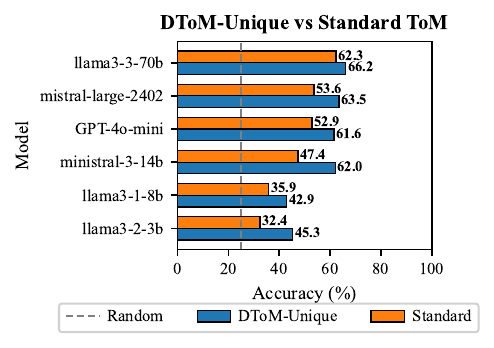}
  \caption{Temporal vs. standard ToM questions.}
  \label{fig:dtom_vs_standard}
\end{subfigure}

\caption{Model performance across question types and evaluation settings.}
\label{fig:three_results}
\end{figure*}

\section{Experimental Results}

\noindent \textbf{Overall Performance.}
Table~\ref{tab:main_results} presents the main results across all question types and models. Overall accuracy ranges from 35.7\% (LLaMA 3.2-3B) to 63.3\% (LLaMA 3.3-70B), with all models performing well above the random baseline of 25\%. Larger models generally outperform smaller ones, although GPT-4o-mini (55.1\%) performs comparably to larger models in several cases. These results indicate that current LLMs support limited forms of dynamic ToM reasoning, while also revealing clear limitations in maintaining and updating beliefs over extended interaction.


\noindent \textbf{Performance by Question Type.}
Fig.~\ref{fig:question_difficulty} show that update-detection questions yield the highest accuracy (67.5\%), indicating that models can often identify when belief changes occur. Post-update questions also show strong performance (63.9\%), suggesting effective inference of current beliefs. In contrast, pre-update questions are markedly more challenging (27.7\%), revealing persistent difficulty in maintaining access to prior mental states after belief revision. Second-order questions exhibit intermediate performance (47.2\%) with false belief questions slightly lower (44.7\%), consistent with the effectiveness of the information-asymmetry design, in which hidden inner speech reliably induces false beliefs.

\noindent \textbf{Recency Bias.} 
Interestingly, we observe a pronounced difference between post-update and pre-update performance (Fig.~\ref{fig:recency_bias}). Averaged across models, accuracy on post-update questions (63.9\%) substantially exceeds that on pre-update questions (27.7\%), indicating a strong preference for reasoning about current rather than prior beliefs. This pattern holds across model families and scales. LLaMA~3.1–8B shows the largest gap between post- and pre-update accuracy, and even the strongest model, LLaMA~3.3–70B, exhibits reduced accuracy when recalling beliefs held before an update. 
Overall, these results suggest that while LLMs can reliably infer updated mental states, they have difficulty maintaining and retrieving earlier belief representations once new information is introduced, consistent with a recency bias in which recently updated beliefs dominate reasoning~\citep{apperly2010mindreaders}.

\noindent \textbf{Comparison with Standard ToM Evaluation for LLMs.} Fig.~\ref{fig:dtom_vs_standard} compares performance on the temporal question types introduced in DToM-Track (pre-update, post-update, update detection) with standard ToM questions, including temporal, second-order, and false-belief reasoning. Across models, temporal questions reveal a consistent difficulty in tracking belief change over time that is not captured by static ToM evaluations. In contrast to existing LLM-based ToM frameworks that focus on inferring mental states at a single point in a conversation~\citep{tomato2024}, DToM-Track exposes the added challenge of recalling beliefs held prior to an update. Specfically, pre-update questions are substantially more difficult than false-belief questions (27.7\% vs.\ 44.7\%), indicating that maintaining belief trajectories over interaction is a distinct and underexplored dimension of ToM.

\section{Discussion and Conclusions}
This work reframes ToM as a temporally extended cognitive process that requires not only inferring others' mental states but also maintaining and retrieving those representations as beliefs evolve over interaction. To operationalize this view, we introduce DToM-Track, a framework that probes dynamic ToM by separating current-belief inference from access to prior belief states after revision, using LLMs as computational probes.
Across six LLMs, we observe a clear dissociation: models reliably infer current belief states but struggle to maintain belief trajectories over time. Accuracy drops sharply for pre-update belief recall, consistent with recency bias and interference, in which recent updates dominate retrieval and obscure earlier mental states. This pattern persists in larger models, suggesting limitations in temporal representation and retrieval rather than model scale alone.

\noindent \textbf{Implications.} 
These findings align with cognitive evidence that belief reasoning over time imposes demands beyond static belief attribution~\citep{hogarth1992order}. While classic false belief tasks capture reasoning about beliefs that conflict with reality at a given moment, maintaining and retrieving belief representations over time is shaped by general-purpose memory mechanisms, including recency bias and retrieval interference. From this perspective, failures in dynamic ToM may reflect difficulties in accessing earlier belief states after updates rather than an inability to represent others' beliefs. DToM-Track provides an empirical framework for isolating these temporal and memory-related components of ToM that are obscured in standard evaluations. By treating LLMs as computational probes, it supports analysis of ToM as a temporally extended process~\citep{wilf2024think} and enables comparison with computational ToM models and cognitive theories that aim to explain and predict belief reasoning over time~\citep{nguyen2024predicting,frank2025cognitive}.

\noindent \textbf{Limitations and Future work.} 
This work provides formative evidence for evaluating dynamic aspects of ToM, with several limitations that motivate future research. First, our analysis focuses on zero-shot performance and does not examine whether fine-tuning could mitigate recency bias and interference. Second, DToM-Track relies on synthetic role-play conversations with information asymmetry and linguistic markers to guide update instructions, which may cue models. 
Future work will examine these effects through ablations that manipulate the presence and strength of update cues. An important next step is to test whether similar recency and interference patterns emerge in human judgments and naturalistic dialogue, and how they scale with interaction length and belief revision~\citep{jones2024}.
Finally, integrating DToM-Track with computational ToM models can support mechanistic accounts of social reasoning in humans and artificial agents.



\printbibliography

\end{document}